\documentclass[runningheads]{llncs}
\usepackage{graphicx}
\usepackage{amsmath,amssymb} 
\usepackage{color}
\usepackage{times}
\usepackage{epsfig}
\usepackage{graphicx}
\usepackage{amsmath}
\usepackage{amssymb}
\usepackage{lipsum}
\usepackage{tabulary}
\usepackage{booktabs}
\usepackage{adjustbox}
\usepackage[width=122mm,left=12mm,paperwidth=146mm,height=193mm,top=12mm,paperheight=217mm]{geometry}

\begin{document}

\pagestyle{headings}
\mainmatter
\def\ECCV18SubNumber{2821}  

\title{Unsupervised Deep Features for Remote Sensing Image Matching via Discriminator Network} 

\titlerunning{Unsupervised Remote Sensing Image Matching}

\authorrunning{Mohbat et. al}

\author{Mohbat Tharani, Numan Khurshid, Murtaza Taj}
\institute{Lahore University of Management Sciences, Lahore, Pakistan}

\maketitle





\begin{abstract}
The advent of deep perceptual networks brought about a paradigm shift in machine vision and image perception. Image apprehension lately carried out by hand-crafted features in the latent space have been replaced by deep features acquired from supervised networks for improved understanding. However, such deep networks require strict supervision with a substantial amount of the labeled data for authentic training process. These methods perform poorly in domains lacking labeled data especially in case of remote sensing image retrieval. Resolving this, we propose an unsupervised encoder-decoder feature for remote sensing image matching (RSIM). Moreover, we replace the conventional distance metrics with a deep discriminator network to identify the similarity of the image pairs. To the best of our knowledge, discriminator network has never been used before for solving RSIM problem. Results have been validated with two publicly available benchmark remote sensing image datasets. The technique has also been investigated for content-based remote sensing image retrieval (CBRSIR); one of the widely used applications of RSIM. Results demonstrate that our technique supersedes the state-of-the-art methods used for unsupervised image matching with mean average precision (mAP) of 81\%, and image retrieval with an overall improvement in mAP score of about 12\%.

\end{abstract}





\section{Introduction}
\label{Introduction}

 

Remote sensing image acquisition technology has been actively playing a vital role in the advancements of geological analysis, weather forecasting, remote resource localization, urban infrastructure and planning, and natural hazard monitoring. Recent satellite image datasets and unsolved image perception problems requiring improved solutions, have successfully gathered the attention of computer vision scientists. In general, high resolution remote sensing image matching solely relies on the efficiency of the methods employed for selection and extraction of image features; discriminated through various standard similarity metrics \cite{FamaoYe2018RSIRCNN}. 
 
\begin{figure*}[t]
    \centering
    \includegraphics[width=0.99\linewidth]{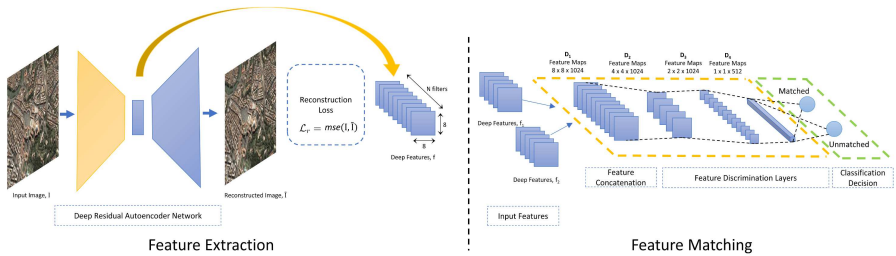}
    \caption{Proposed architecture for Remote Sensing Image Matching, where the section on the left depicts the process of extracting unsupervised features of the remote sensing image while the section on the right shows the discrimination process of the image feature pair.}
    \label{fig:pipeline}
\end{figure*}
Classical image matching techniques involve various shallow hand crafted visual descriptor to represent the content of the matching images. Colours, edges, corners, spatial histogram, and morphological features \cite{Bosilj2016} spread across the image are regarded as global hand-crafted features while scale invariant feature transform (SIFT), Bag-of-visual-words (BoVW) \cite{Aptoula2014} and Bag-of-Features (BoF) \cite{Passalis2017LearningNB} techniques profess to extract local hand-crafted features of the image for representation in lower dimensional space.

Usually regarded as the combination of the global and local features~\cite{Chen2017b}, learned features on the other hand are not calculated by predefined algorithms, instead they are inferred through the cognition of convolutional neural network (CNN) architectures. The class-learnable architecture may consist of several processing layers containing up-to thousands of weight filters, coupled with non-linear operators, learned in an end-to-end manner~\cite{drozdzal2016skipconnections}. CNN networks are generally trained for the image classification via back-propagation on the given set of training images with their corresponding class labels and are then used to predict the label of test image set. The fully connected layer of the trained network is usually taken as features of the given image ~\cite{SharifRazavian2014} which has also been employed previously for remote sensing image retrieval \cite{FamaoYe2018RSIRCNN}. Despite the fact that these networks perform extraordinary for image classification, they have a potential drawback of need for the enormous labeled images which are very difficult to collect and analyze. Putting it technically, supervised CNN in remote sensing environment usually suffer from either insufficient training samples or lack of balanced datasets. Moreover, their vector-based feature alignment causes loss of structural information present in satellite imagery effecting the computation of overall discrimination between the image pair \cite{LichaoM2017UnsupResFeat}.

Recently, autoencoder based unsupervised methods for feature extraction have been used for classification of diverse set of images including remote sensing images~\cite{VincentPASCALVINCENT2010,LichaoM2017UnsupResFeat}. An improved way of using such autoencoder features is to exploit the residual blocks being used in the network which increases its efficiency substantially~\cite{LichaoM2017UnsupResFeat,Chen2017a} through faster convergence, solving vanishing gradients problem, and retaining significant information of the image by introducing shorter paths with fewer non-linearities to deep layers of the network \cite{drozdzal2016skipconnections}. 

Traditional similarity metrics e.g. Euclidean distance or Cosine similarity computed for image feature pair fully depend on the pixel level information in the feature vector, considering low-level patterns, found in matching features, inconsequential~\cite{Larsen2015}. Instead, distance metric learning approaches gained significant popularity, which actually replaces conventional metrics by learning the the global or local patterns of the features to discriminate~\cite{WanDL4CBIR,hoi2006globalsuperlearning}. Based upon a loss function, minimized to learn the disparities between the images, recently introduced deep discriminative networks learns a combination of local and global pattern and outputs a probability of its success to differentiate the image features. Siamese~\cite{Koch2015SiameseNN} and Triplet networks~\cite{HofferTripletNetwork2015} for example, have replaced ordinary numerical similarity measures with learned discriminating layer, used in variety of applications e.g. Crossview matching between street-view and satellite-view images~\cite{Vo2016}. Apart, a CNN based well known discriminative network has been used by Goodfellow et. al. in Generative Adversarial Networks (GANs) and its variants \cite{Goodfellow2014}, while~\cite{Larsen2015} used the same for face discrimination.

We present a novel framework combining the salient features of unsupervised visual descriptors with a novel deep discrimination process for image matching described in Fig.~\ref{fig:pipeline}. The whole process could be divided into two main phases: in feature extraction step deep features of the image pair would be extracted while in discrimination step these features would be used to train a deep residual discriminator network for decision making. The performance of this image matching framework is then tested with remote sensing image retrieval problem considering its fundamental steps. Our contributions in this research could be listed as follow:

\begin{itemize}
  \item We developed a residual encoder-decoder based visual descriptor to be used for remote sensing image matching and retrieval showing that autoencoder based unsupervised features perform considerably better as compare to other unsupervised features.
  \item We propose to replace conventional similarity metrics with a learned model for feature matching by designing a residual discriminative network, for discriminating the features more efficiently.
  \item A comparative evaluation of various classifiers for the extracted unsupervised features has been illustrated.
  \item We demonstrated the effectiveness of our model with two large benchmark remote-sensing datasets, containing roughly over $1.5\times10^{5}$ sq/km of area.
\end{itemize}

The remainder of this paper is organized as follows. In Section \ref{ProposedMethod} the RSIM framework is described. In Section \ref{Experimentation} the experimental design and analysis is provided while results have been presented in Section \ref{Results}. The paper has been concluded in Section \ref{Conclusion}.

{}

\begin{figure*}[t]
    \centering
    \begin{tabular}{c}
        \includegraphics[width=0.9\textwidth]{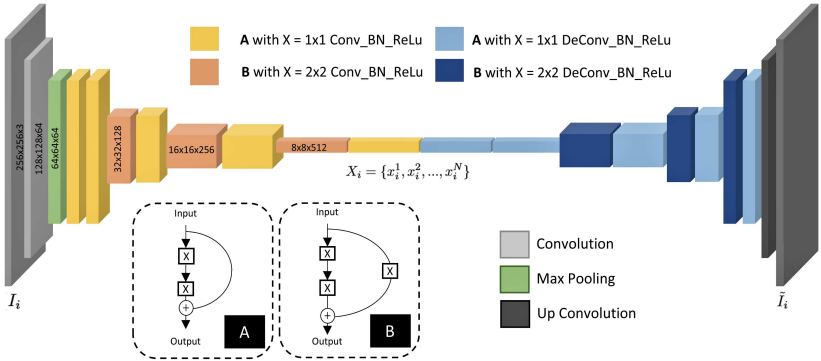}\\
        
    \end{tabular}
    
    \caption{Unsupervised Autoencoder Features: Image input from left and outputs to the right of network. Two type of residual blocks mentioned as A and B have been used. The color codes for each layer define the use of each block with specific combination of layers specified by sub-block X in the residual block. $X_i$ is taken as the feature vector of the given image}
    \label{fig:autoencoder_full}
\end{figure*}

\section{Proposed Method}
\label{ProposedMethod}

To provide solution to the issues discussed in the previous section, we propose a novel architecture for RSIM system involving multi-tier residual networks for learning optimal visual descriptors of the images and a deep discriminative network for differentiating the query image features from its matching image features. The performance of the architecture would then be evaluated for the problem of  CBRSIR on benchmark datasets.

\subsection{Image Matching}
The idea is to exploit the cognition and accuracy of residual layer based deep auto-encoders to learn powerful features 
of all the images in dataset, $\mathcal{D}$. Consider a large dataset,  $\mathcal{D}$, containing $T$ number of total remote-sensing images can be given as $\mathcal{D}={\{\mathbf{I}_{1},\mathbf{I}_{2},...,\mathbf{I}_{T}\}}$ where $\mathbf{I_i}$
represents an image. Latent features of $i^{th}$ image, $\mathbf{I_{i}}$, can be described as ${{X}_{i}=\{x_{i}^{1},x_{i}^{2},...,x_{i}^{N}\}}$ where $N$ is the total number of features of each $i^{th}$ image. 
Let, we have two images $I_a$ and $I_b$, making image pair $(I_a,I_b)$ would consist of the concatenated features  $(X_a,X_b)$ resulting features $(\{x_{a}^{1},x_{a}^{2},...,x_{a}^{N}\},\{x_{b}^{1},x_{b}^{2},...,x_{b}^{N}\})$. This pair of features is then passed through a network to compute a probability of similarity between them. Graphical illustration of the proposed architecture is shown in Fig.~\ref{fig:pipeline}. Left side of the figure shows the feature learning network and feature extraction methodology for the images while right side of the figure describes the feature matching block used to discriminate in pair of image features. The details of each block will be described in Section \ref{DeepRes Features} and \ref{FeatDisc} of this paper.

\subsection{Image Retrieval}
Content based remote sensing image retrieval, one of the very well known applications of the RSIM, normally employs supervised features acquired through pre-trained deep learning models for image matching purpose. In RSIM, the query image $I_{q}$, having features  $\{x_{q}^{1},x_{q}^{2},...,x_{q}^{N}\}$ may belong to either test set $\mathcal{D}$, i.e. $I_{q} \in \mathcal{D}$ or it may be taken from somewhere else as unseen sample i.e. $I_{q} \notin{\mathcal{D}} $. As discussed in  \cite{FamaoYe2018RSIRCNN}, traditional CBRSIR systems comprise of indexing, which is applied to all the images of dataset $\mathcal{D}$ as well as the query image, $\mathbf{I}_{q}$. Then, in the retrieval step a subset of images, $\mathcal{D}_{ret}$ is retrieved on the basis of similarity score as $\mathcal{D}_{ret} \subset \mathcal{D}$ having both relevant images $\mathcal{D}_{rel}$ and non relevant images $\mathcal{D}_{nrel}$. Relevance feedback that feeds back the selected relevant images as query for several times and refines the $\mathcal{D}_{ret}$ set by various methods described in \cite{Napoletano2018}, tries to achieve a state where $\mathcal{D}_{rel} = \mathcal{D}_{ret} \cap \mathcal{D}_{rel}$.
Moreover, manual quantitative evaluation performed by human observer through visualization is sometime used to re-analyze $D_{ret}$ to identify relevant images, $D_{rel}$, in $D_{ret}$, which is considered to be one of the tedious and time consuming process.
In our method, deep residual features acquired through autoencoder for matching image ${X_{m}} \in \mathcal{D}$ and query image, $X_{q}$, are fed to another residual CNN matching network which identifies either query image and matching image are similar with certain probability or not. Afterwords, top $n$ images out of retrieved image-set $\mathcal{D}_{ret}$ ordered according to their matching probability and referred as relevant images, $\mathcal{D}_{rel}$.

\subsection{Unsupervised Autoencoder Features}
\label{DeepRes Features}
Inspired by \cite{LichaoM2017UnsupResFeat}, our feature extraction method is a deep residual CNN based autoencoder consist of alternately stacked residual layers having convolution and maxpooling layers in the encoding block while de-convolutional layers in the decoding block. The middle layer of the network is considered to be the layer of interest employed as discriminative features, as illustrated in Fig.~\ref{fig:autoencoder_full}. This feature representation might not be very compact as compare to the one acquired in supervised method, however, it retains the structural information of the image and is able to reconstruct the original image using a decoder network. Encoder outputs the best representation of input image with only $15\%$ of the original data which is an efficient dimensionality reduction solution. Moreover, the network does not involve any fully connected layers resulting in reduced set of learning parameters. In our case the network has been trained on the $80\%$ of the data and has been tested with the rest of $20\%$ from each of the benchmark datasets. As a result, intrinsic features of the input images have been learned by transforming them to latent space. The decoder transforms the deep feature back to image space and mean squared error loss,$\mathcal{L}_{r}$ is computed between the input image and the decoded image. Once the network has been fully trained, it has been used to develop a feature database for all the images of the dataset. The overall system can be represented by the following sequential equations.
\begin{equation} \label{eq:enc}
X_i=enc(I_i,\theta_e),
\end{equation}
\begin{equation} \label{eq:features}
X_{i}=\{x_{i}^{1},x_{i}^{2},...,x_{i}^{N}\},
\end{equation}
\begin{equation} \label{eq:dec}
\tilde{I_i}=dec(X_i,\theta_d),
\end{equation}
\begin{equation} \label{eq:loss1}
\mathcal{L}_{r}=\dfrac{\sum_{j=1}^{n}{(I_i^j-\tilde{I_{i}^{j}})}^{2}}{n},
\end{equation}
%
%
where $\theta_e$ and $\theta_d$ are hyperparameters of the encoding and decoding parts of the autoencoder network, respectively and $n$ is the total number of pixels in image $I$. 

\begin{figure}[t]
    \centering
    \includegraphics[width=0.7\columnwidth]{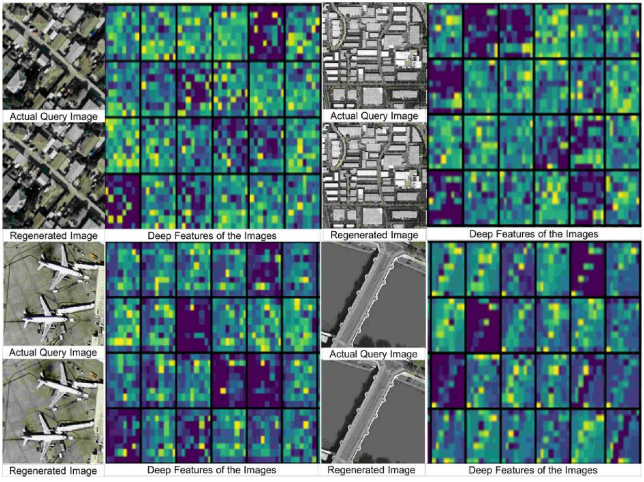}
    \caption{Autoencoder representative features shown with its actual and regenerated image. Out of $512$ features, we have visualized only $24$ features each of size $8\times 8$. It can be seen that these features maintain the structural information while encoding the image}
    \label{fig:ResultsAutoencoder}
\end{figure}

\subsection{Feature Discrimination}
\label{FeatDisc}
Standard RSIM and CBRSIR systems use similarity metrics to discriminate the set of image features. In our proposed discrimination process the features from image pair are concatenated and passed through a series of convolution layers until they are fed to a dense layer with a softmax activation which classifies them to a match or a mismatch as shown in Fig.~\ref{fig:pipeline}. The weights of the layers are trained on the training set of images before using it as a discriminator to identify the matching image pair. The same network if used in remote sensing image retrieval would train to identify the match between the query image features and the target image features, consequently replacing the need to use tedious and time consuming relevance feedback operation. 
For the given query image, $I_q$ having unsupervised autoencoder features, $X_q$ and the matching feature $X_{m}$ from the database, the discriminative network can be described with the following set of equations:
\begin{equation} \label{eq:disc1}
\tilde{y}=disc(enc(I_q,\theta_e),enc(I_m,\theta_e),\theta_c),
\end{equation}
\begin{equation} \label{eq:disc2}
\tilde{y}=disc((X_q,X_m),\theta_c),
\end{equation}
where $\tilde{y}$ is the predicted label by the network and $\theta_c$ describes the hyper-parameters of the discriminative network. For this network the binary cross-entropy loss could be formulated as: 
\begin{equation} \label{eq:loss2}
\begin{aligned}
\mathcal{L}_{c}=-\sum_{q,m} y\log \tilde{y},
\end{aligned}
\end{equation}
where $y$ is the true label and $\forall(q,m), y \in (0,1)$, where $1$ if the input images are from the same class (similar), while $0$ if input query and target images belong to different class.

\begin{figure}[t]
    \centering
    \includegraphics[width=0.9\linewidth]{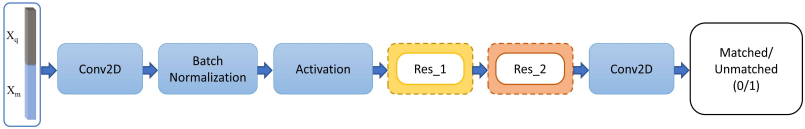}
    \caption{Architecture of the proposed discriminator network that takes deep autoencoder features of an image pair and predicts the matching score}
    \label{fig:DeepRetNet}
\end{figure}

Stacking the features of the query image and sample image in the third dimension allowed us to have an array of resolution $8\times8$ as input to our discriminative network. Such diminished resolution allowed us to design our network in such a way that it emphasize on learning the depth dimension with the help of additional batch normalization and ReLU layers. Apart, to learn the  disparities in the stacked features, two types of residual blocks with different strides have been used to train the network. The label is predicted by applying a dense layer on the last flattened convolution layer.

\begin{figure*}[t]
    \centering
    \includegraphics[width=0.99\linewidth]{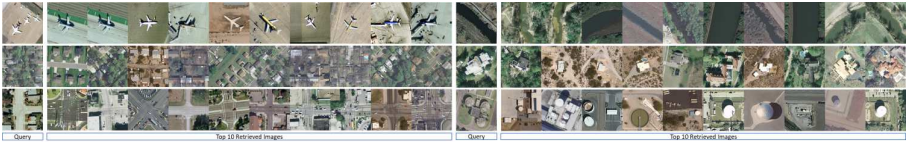}
    \caption{Top 10 retrieved images for the corresponding query images of LandUse Dataset. The retrieval results shown for queries taken from the classes of Airplane, Dense Residential, Intersection \textit{(Left)},  River, Sparse Residential, and Storage Tanks \textit{(Right)}.}
    \label{fig:retrievalResults}
\end{figure*}

Simplicity of the discriminative network resulting in fewer number of learnable filters enabled the network to gradually converge towards accurate prediction of labels. The network is concluded with a softmax activation with binary cross entropy loss,$\mathcal{L}_{c}$ (see equation ~\ref{eq:loss2}) for match prediction. The detailed architecture has been depicted in Fig.~\ref{fig:DeepRetNet}. Similar optimization, learning rate and early stopping criteria have been maintained as used while optimizing autoencoder network previously.

\section{Experimentation and Analysis}
\label{Experimentation}

We evaluated our approach on two benchmark remote sensing datasets namely University of California Merced Land Use/Land Cover dataset (LandUse) and High-resolution Satellite Scene dataset (SceneSet). The LandUse Dataset contains 21 diverse classes with aerial orthoimagery of 1 feet per pixel resolution. Each class contains $100$, $256\times256$ images covering total area of about $4.2\times{10}^4$ sq/km in $2100$ images. Some of its examples are shown in Fig.~\ref{fig:ResultsAutoencoder}
The satellite scene dataset (SatScene) contains 19 diverse classes with aerial orthoimagery of different zoom level. Each class contains 100, $600\times600$, $1005$ images in total. Some of its examples are shown in Fig.~\ref{fig:ResultsAutoencoder}
It is quite evident from the images in figure that both datasets are quite different in terms of pixel resolution, sizes, zoom levels, and classes making the scenario more challenging for our proposed learning model.

We trained our autoencoder in an unsupervised manner (see Sec.~\ref{DeepRes Features}) on $200$K Google street-view images and $200$K satellite-view images of GTCrossView dataset~\cite{Vo2016} and then fine-tuned it on the datasets used in this study (LandUse and SatScene). The training converges at MSE loss of approximately $250$, depicting a very small difference in input and regenerated image of autoencoder.

\subsection{Performance Metrics}
\label{performanceMetrics}
Widely adopted by the remote sensing community the performance of image matching has been evaluated by Mean average precision (mAP), while retrieval systems employ average normalized modified retrieval rank (ANMRR), and class-wise mAP as performance indicators. \cite{Ozkan2014,Napoletano2018}. Major advantage of ANMRR is that it considers the number of similar images that are retrieved and quantifies them as per their rank which also address the queries having varying relevant image sets in image retrieval problem. Mathematically Rank(\textit{k}) is defines as:
\begin{equation}\label{rank3}
\begin{aligned}
Rank(k)&=\begin{cases}
  Rank(k), \quad if Rank(k)\le k(q) \\  
  1.25K(q), \quad if Rank(k)> K(q)   
\end{cases}
\end{aligned}
\end{equation}
where \textit{Rank(k)} is the \textit{k}th position at which a similar item is retrieved.

\textit{G(q)} is the set of relevant images. \textit{K(q)} is constant penalty and is commonly chosen to be \textit{2G(q)}. $Rank_{mean}(\textit{q})$ is defined as:
\begin{equation}\label{rank2}
\begin{aligned}
{Rank_{mean}(q)=\frac{1}{G(q)}\sum_{k=1}^{G(q)}Rank(k)}
\end{aligned}
\end{equation}
Normalized modified retrieval rank (NMRR) is described as:
\begin{equation}\label{rank0}
\begin{aligned}{NMRR=\frac{[Rank_{mean}(q)-0.5[1+G(q)]}{1.25K(q)-0.5[1+G(q)]}}
\end{aligned}
\end{equation}
For which average NMRR can then be calculated as:
\begin{equation}\label{rank1}
\begin{aligned}{ANMRR=\frac{1}{Q}\sum_{q=1}^{Q}NMRR}
\end{aligned}
\end{equation}
where \textit{Q} indicates the number of queries \textit{q} performed.

\begin{figure*}[t]
    \centering
		\includegraphics[width=0.99\linewidth]{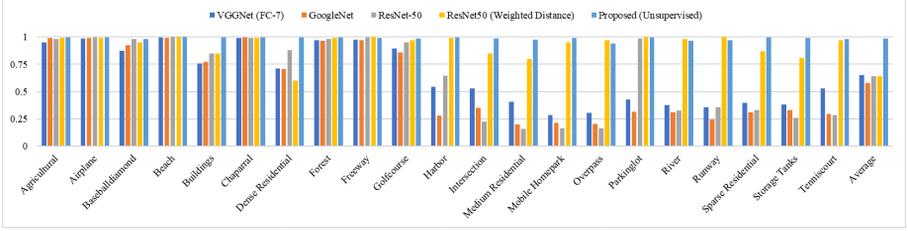}
    \caption{Comparative results of Class-wise mAP among VGGNet, GoogleNet, ResNet-50, ResNet-50 with Weighted Distance \textit{(Supervised)}, and our proposed approach \textit{(Unsupervised)}. Our approach surpasses the efficiency of supervised techniques in average mAP computed for all the classes of LandUse Dataset.}
    \label{fig:ClassWiseMAP}
\end{figure*}
 \setlength{\tabcolsep}{4pt}
\begin{table}[b]
\begin{center}
\caption{A comparison of Mean Average Precision (mAP) values among traditional similarity metrics and our discriminator network applied to unsupervised autoencoder features for LandUse Dataset. The discriminator network is trained with 80\% of the training data and tested with the rest of the 20\%.}
\label{Table1:ClassAcc}
\begin{tabulary}{\linewidth}{CCC}
\toprule
\textbf{Discrimination Measure  }&\textbf{mAP}\\
\midrule

Manhattan Distance& 6.286\\
Euclidean Distance& 4.644\\
Cosine Similarity& 4.789 \\
Softmax Classifier& 50.523\\
Proposed Discriminator Network& 81.201 \\

\bottomrule
\end{tabulary}
\end{center}
\end{table}
\setlength{\tabcolsep}{1.4pt}

\setlength{\tabcolsep}{4pt}
\begin{table*}[t]
\begin{center}
\caption{Comparative evaluation of our proposed approach with the state-of-the-art hand-crafted and supervised techniques. It should be noted that our proposed approach does not use any Relevance Feedback Mechanism (instead we adopted Basic Retrieval System).}
\begin{adjustbox}{width=12cm}
\begin{tabular}{lrrrrrrrrr}
\hline\noalign{\smallskip}
Features &Feature Type & ANMRR & mAP & P@5 & P@10 & P@50 & P@100 & P@1000\\ 

\hline

\bf LandUse Dataset &&&&&&&&\\

HoG  \cite{Napoletano2018}& Hand-Crafted & 0.751 & 17.85& 48.67& 41.88& 25.37&19.2 &6.18\\ 
LBP RGB \cite{Napoletano2018}&Hand-Crafted & 0.751 & 17.96& 58.73& 49.83& 28.12&19.62&6.07\\ 
Dense SIFT (VLAD) \cite{zhou2017learning}&Hand-Crafted&0.649  &28.01& 74.93& 65.25& 38.20&28.10 &7.18 \\ 
Dense SIFT (FV) \cite{Napoletano2018}&Hand-Crafted& 0.639 & 29.18&   75.34 & 66.28 & 39.09 & 28.54 & 7.88 \\ 
GoogleNet \cite{szegedy2015going}& Deep-Supervised & 0.360 & 55.86& 85.36& 80.96& 64.71&52.36 &9.68\\ 
NetVLAD \cite{arandjelovic2016netvlad}&Deep-Supervised& 0.406 &  51.44& 83.00& 78.59&61.63 &49.04&9.29 \\ 
MLIR CNN-Fc7 \cite{shao2018MLIR}  &Deep-Supervised& 0.322 & 62.73&   80.76 & 71.00 & 30.80 & 17.77 & - \\ 
SatResNet-50 \cite{Napoletano2018}  &Deep-Supervised& 0.239 & 69.94&   92.06 & 89.02 & 77.23 & 64.42 & 9.86 \\ 
Discriminator NW [Proposed] & Deep-Unsupervised & \bf 0.09&\bf 81.20 & \bf 100&\bf 99.2& \bf 99.2& \bf 87.4& \bf 9.90\\ 
\hline
\bf SatScene Dataset &&&&&&&&\\

HoG  \cite{Napoletano2018}& Hand-Crafted &  0.724& 19.97& 40.24 &35.31 &21.73 &15.82 &5.20 \\
LBP RGB \cite{Napoletano2018}&Hand-Crafted &  0.664& 24.95 &50.33& 43.98& 26.33& 19.40& 5.20 \\
Dense SIFT (VLAD) \cite{zhou2017learning}&Hand-Crafted&0.649  &28.01& 74.93& 65.25& 38.20&28.10 &7.18 \\
Dense SIFT (FV) \cite{Napoletano2018}&Hand-Crafted& 0.552& 35.89 &71.30& 62.78& 36.19 &25.03 &5.20  \\
GoogleNet \cite{szegedy2015going}& Deep-Supervised &  0.324& 60.36 &85.73 &82.28 &68.32 &55.75& 9.75 \\
NetVLAD \cite{arandjelovic2016netvlad}&Deep-Supervised&  0.371& 56.37& 82.54 &78.41& 64.40& 52.19 &9.48  \\
SatResNet-50 \cite{Napoletano2018}  &Deep-Supervised&  0.207& 74.19& 92.11& 90.55 &80.91&\bf 68.02 &9.87 \\
Discriminator NW [Proposed] & Deep-Unsupervised & \bf 0.06&\bf96.6 & \bf 100&\bf 100& \bf 94.3&  52.00& \bf 9.92\\

\hline
\end{tabular}
\end{adjustbox}
\end{center}
\label{Table2:CBIR}
\end{table*}
\setlength{\tabcolsep}{1.4pt}

Precision on the other hand could be defined as the fraction of retrieved images relevant to query image. It is usually evaluated in the cut-off rank, considering topmost k results yielded by CBRSIR system. This measure is termed as P@k. In this research we are calculating Mean Average Precision (MAP) and per class MAP values for the comparison with state-of-the-art methods. Mathematically, mAP can be computed as:

\begin{equation}\label{maP4}
\begin{aligned}
{mAP=\frac{1}{Q}\sum_{q=1}^{Q}AP(q)}
\end{aligned}
\end{equation}
where average precision (AP) is:
\begin{equation}\label{maP5}
\begin{aligned}
AP=\frac{\sum_{k=1}^{n}(Precision(k)\times z(k))}{Number\, of\, relevant\, images}
\end{aligned}
\end{equation}
where $n$ is the number of retrieved images, $k$ is their rank, and $z(k) \in (0,1)$, equaling $1$ if the feature at rank $k$ belong to a relevant image, while zero otherwise.

\section{Results and Discussion}
\label{Results}

Residual blocks constituting autoencoder network avoid vanishing gradient problem resulting in less overfitting and hence better optimization. Our network generalizes to efficiently encode images of multiple disciplines, including but not limited to street view, satellite view, medical imagery, and synthetically generated images, into low dimensional space. Some of the learned deep features are visualized in Fig.~\ref{fig:ResultsAutoencoder}, in which the triplets contain a query image, its deep features and a decoded image.

\subsection{Remote Sensing Image Matching}
The performance of discriminator network used to distinguish between similar and dissimilar image pair is compared with the rest of traditional similarity metrics for unsupervised autoencoder features in Table \ref{Table1:ClassAcc}. It has been clearly observed that these conventional metrics i.e. Euclidean Distance and Cosine Similarity fail to perform for our proposed unsupervised autoencoder features. Even when softmax classifier is unable to learn the disparity between the unsupervised visual discriptors, our proposed discriminator network successfully finds a non-linear decision boundary between these features, outperforming other techniques by achieving mAP of 81.2\% for LandUse Dataset.






\subsection{Remote Sensing Image Retrieval}
We showed that learnt discriminative network replacing numerical similarity metric calculation employed with unsupervised features works far more efficiently as compare to other supervised/unsupervised CBRSIR approaches. We evaluated our approach on two benchmark dataset on 20\% split of test images from both datasets for the performance metrics discussed in the section \ref{performanceMetrics}. In comparison with previous approaches using hand-crafted and CNN based supervised features our approach performs well interm of ANMRR, precision and mAP. Top 5 retrieved images always belong to the class of the query image while top $10$ images have also been recognized with $99.2\%$ precision in LandSet and $99.8\%$ precision in SatScene dataset as shown in Table \ref{Table2:CBIR}. Even with top $50$ images our proposed technique left all other approaches way behind. There is a significant improvement in retrieval of top $100$ images which is almost $23\%$ better than the best performing ($64.42\%$) schemes and features used in literature~\cite{Napoletano2018} for basic retrieval method. Unlike SatResNet-50 and Multi-label image retrieval (MLIR)  which fails on classes with objects rotated and translated on image plane (e.g. intersection, dense residential, sparse residential and storage tanks), our model generalized learning is robust to such changes and performs comparable to other classes. Therefore, the average mAP value of our proposed approach is much higher as compare to existing techniques as shown in Fig.~\ref{fig:ClassWiseMAP}. Top 10 retrieval results for their corresponding query images of some of the classes with significant mAP is shown in Fig.~\ref{fig:retrievalResults}. 

\begin{figure*}[t]
    \centering
    \includegraphics[width=0.8\linewidth]{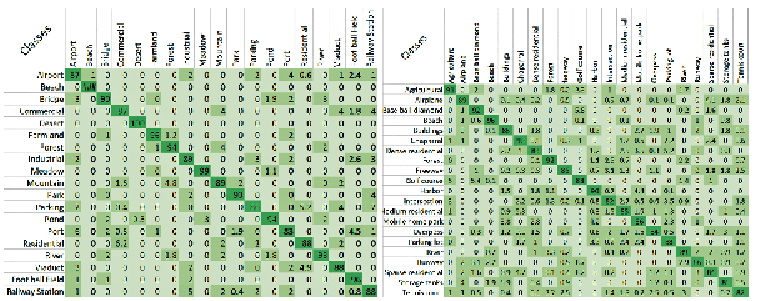}
    \caption{Confusion metrics of LandUse and SatScene dataset averaged over $10$ queries per class}
    \label{fig:charts1}
\end{figure*}
Class-wise percentage accuracy is another important parameter, clearly representing the overall performance of our approach. For this metric, we computed the measure of relevance of the given query image with $n$ retrieved images identified by our proposed framework ($n=100$, if the given class contain $100$ images only). The relevance of each retrieved image is associated with its actual class. The process is repeated for 10 queries from the same class and the averaged results are shown in a confusion matrix given in Fig.~\ref{fig:charts1} effecting the overall performance of the system. It can be observed that images from aeroplane class were confused with images from storage tanks class, while tennis court images and freeway images have also been confused due to similar visual features.

\section{Conclusion}
\label{Conclusion}
In this paper, we addressed the scarcity of labeled remote sensing data by generalizing existing RSIM and CBRSIR systems, firstly by introducing deep unsupervised features and secondly with a novel discrimination method replacing crusted similarity measures. We proposed a novel autoencoder architecture and then used the features from its middle layer as our visual descriptors for image matching. We demonstrated that while these features are compact in terms of representation, they can be used to discriminate images in a better way. Unlike, existing literature of CBIR for remote sensing, we for the first time replaced the conventional distance measure with a discriminative network that uses the autoencoder features as its input and discriminates between features of relevant and irrelevant images. Evaluation of two benchmark datasets shows that our proposed framework outperforms the existing literature and accomplishes to achieve far better ANMRR and precision values. Furthermore, our approach is easy to train as we completely eliminate the need for iterative and tedious relevance feedback step that require user-provided annotations which are hard to find for Remote Sensing data.









\bibliographystyle{splncs}


%



\end{document}